\documentclass[runningheads]{llncs}

\usepackage{eccv}

\usepackage{eccvabbrv}

\usepackage[utf8]{inputenc} 
\usepackage[T1]{fontenc}    
\usepackage{url}            
\usepackage{booktabs}       
\usepackage{amsfonts}       
\usepackage{nicefrac}       
\usepackage{microtype}      

\usepackage{graphicx}
\usepackage{amsmath}
\usepackage{amssymb}

\usepackage{verbatimbox}
\usepackage{diagbox}
\usepackage{multicol}
\usepackage{enumerate}
\usepackage{epsfig}
\usepackage{threeparttable}
\usepackage{enumitem}
\usepackage{multirow}
\usepackage{color}
\usepackage{array}
\usepackage{setspace}
\usepackage{makecell}
\usepackage{wrapfig}
\usepackage[accsupp]{axessibility} 

\usepackage[pagebackref,breaklinks,colorlinks]{hyperref}
\usepackage{orcidlink}

\begin{document}
	
	\title{Cog2Gen3D: Sculpturing 3D Semantic-Geometric Cognition for 3D Generation} 
	
	\titlerunning{Cog2Gen3D}
	
	\author{Haonan Wang\inst{1} \and
		Hanyu Zhou\inst{2}\thanks{Corresponding author.} \and
		Haoyue Liu\inst{1} \and
		Tao Gu\inst{3} \and
		Luxin Yan\inst{1}}
	
	\authorrunning{H. Wang et al.}
	
	\institute{School of Artificial and Automation, Huazhong University of Science and Technology \and
		School of Computing, National University of Singapore \and
		School of Computing, Macquarie University \\
		\email{whn\_aurora@hust.edu.cn, hy.zhou@nus.edu.sg}
	}
	
	\maketitle

	\begin{abstract}
		Generative models have achieved success in producing semantically plausible 2D images, but it remains challenging in 3D generation due to the absence of spatial geometry constraints. Typically, existing methods utilize geometric features as conditions to enhance spatial awareness. However, these methods can only model relative relationships and are prone to scale inconsistency of absolute geometry. Thus, we argue that semantic information and absolute geometry empower 3D cognition, thereby enabling controllable 3D generation for the physical world. In this work, we propose Cog2Gen3D, a 3D cognition-guided diffusion framework for 3D generation. Our model is guided by three key designs: 1) \textbf{Cognitive Feature Embeddings.} We encode different modalities into semantic and geometric representations and further extract logical representations. 2) \textbf{3D Latent Cognition Graph.} We structure different representations into dual-stream semantic-geometric graphs and fuse them via common-based cross-attention to obtain a 3D cognition graph. 3) \textbf{Cognition-Guided Latent Diffusion.} We leverage the fused 3D cognition graph as the condition to guide the latent diffusion process for 3D Gaussian generation. Under this unified framework, the 3D cognition graph ensures the physical plausibility and structural rationality of 3D generation. Moreover, we construct a validation subset based on the Marble World Labs. Extensive experiments demonstrate that our Cog2Gen3D significantly outperforms existing methods in both semantic fidelity and geometric plausibility.
	\end{abstract}

	\section{Introduction}
	\label{sec:intro}
	
	\begin{figure}[t]
		\centering
		\setlength{\abovecaptionskip}{0.1cm} 
		\setlength{\belowcaptionskip}{-0.7cm} 
		\includegraphics[scale=0.34]{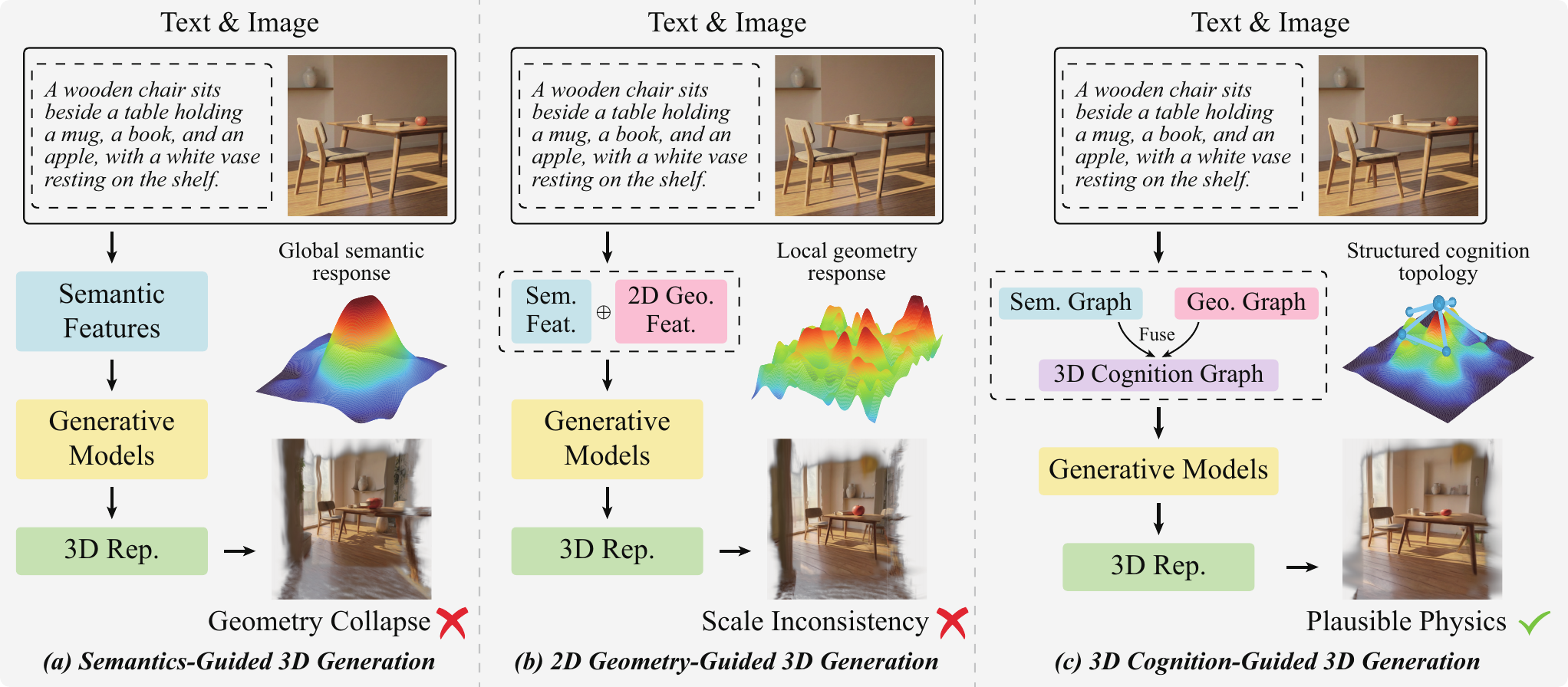}
		\caption{\textbf{Paradigm shift of 3D generation.} Semantic-Guided Generation relies heavily on 2D semantic priors, resulting in physical violations such as object intersections. 2D Geometry-Guided 3D Generation introduces relative geometric constraints but frequently suffers from scale inconsistency due to the lack of absolute metric awareness. Our 3D Cognition-Guided paradigm leverages a 3D cognition graph as the condition, ensuring 3D generation with fidelity and plausibility.}
		\label{fig1}
	\end{figure}

	Generative models have achieved success in synthesizing semantically plausible 2D images, which are widely used in image editing \cite{zhang2023adding} and video generation \cite{blattmann2023stable}.
	When applying these models to the physical world, some researchers rely on the semantic priors of 2D diffusion models and extend them to 3D generation through Score Distillation Sampling \cite{poole2022dreamfusion,wang2023score}, as illustrated in \cref{fig1} (a). However, this paradigm is prone to structural collapse due to the lack of geometric constraints required by the physical world. In this work, our purpose is to integrate geometric and semantic representations to jointly achieve high-quality 3D generation.

	As shown in \cref{fig1} (b), existing methods mainly incorporate geometric priors such as scene graphs \cite{wang2019planit,dhamo2021graph,zhai2023commonscenes,lin2024instructscene,zhai2024echoscene} and layouts \cite{li2019grains,bahmani2023cc3d,po2024compositional,yang2024scenecraft,chen2025layout2scene} to enhance spatial awareness in the 3D generation process. For example, EchoScene \cite{zhai2024echoscene} utilizes scene graphs to define relational dependencies between objects, while Layout2Scene \cite{chen2025layout2scene} relies on bounding box layouts to guide the coarse spatial arrangement of the scene. However, these approaches are limited to modeling 2D relative spatial relationships and fail to capture 3D absolute geometry. This leads to scale inconsistency and geometric collapse, making it difficult to satisfy the rigid constraints of the physical world. Therefore, we argue that the key to 3D generation for the physical world lies in the integration of high-level semantics and absolute geometry.
	
	To address the above issues, we propose that high-level semantics and 3D absolute geometry jointly empower 3D cognition and enhance the physical spatial awareness of 3D generation, as illustrated in \cref{fig1} (c).
	Our design is grounded in two key insights: 1) Geometric features provide \emph{essential physical plausibility}. We figure out that geometric encoders inherently capture dense correspondence and absolute metric information, which motivates us to leverage these cues to enforce strict physical constraints and precise scale consistency within the generative process. 2) Latent scene graphs provide \emph{structural rationality} and \emph{superior reasoning robustness}. We discover that latent scene graphs robustly infer topological relationships in a latent embedding space, which inspires the design of our graph module to ensure the coherent structure of the generated scenes.
	Thus, these observations motivate us to design a common-based fusion scheme that integrates semantic knowledge and geometric information to represent 3D cognition, thereby enabling controllable 3D generation for the physical world.
	
	In this work, we propose \textbf{Cog2Gen3D}, a novel 3D cognition-guided diffusion framework for 3D generation. Our model consists of three key components: 1) \textbf{Cognitive Feature Embeddings}. We encode different input modalities into semantic and geometric representations and further extract logical representation to serve as high-level guidance. 2) \textbf{3D Latent Cognition Graph}. We structure these representations into dual-stream semantic-geometric graphs and utilize the logical representation as a bridge to fuse them via common-based cross-attention, obtaining a unified 3D cognition graph. 3) \textbf{Cognition-Guided Latent Diffusion}. We leverage the 3D cognition graph as condition to guide the diffusion process for 3D Gaussian generation. Through this design, our approach effectively awakens 3D cognition, ensuring that the generated scenes possess both high-fidelity semantics and plausible geometry. Our main contributions are summarized as follows:

	\begin{itemize}
		\item We propose \textbf{Cog2Gen3D}, an innovative framework that introduces 3D cognition to guide 3D generation, effectively bridging semantic priors with geometric constraints to support versatile controllable 3D object and scene generation from arbitrary combinations of visual and textual prompts.
		\item We observe that geometric features provide geometry consistency and latent scene graphs offer structural rationality. This motivates the design of cognitive feature embeddings and 3D latent cognition graph for sculpturing a robust 3D representation that captures appearance attributes and spatial interactions.
		\item We develop a cognition-guided latent diffusion mechanism that steers 3D Gaussian generation using the cognition graph, ensuring both semantic fidelity and geometric plausibility of the generated 3D scenes.
		\item We integrate a series of 3D datasets and construct a curated validation subset, collectively named the cognition scene graph 3D dataset (\textbf{CogSG-3D}), along with explicit scene graph labels for supervision. Extensive experiments have demonstrated the significant effectiveness of our method.
	\end{itemize}

	\section{Related Works}
	
	\begin{figure}[t]
		\centering
		\setlength{\abovecaptionskip}{0.1cm} 
		\setlength{\belowcaptionskip}{-0.7cm} 
		\includegraphics[scale=0.265]{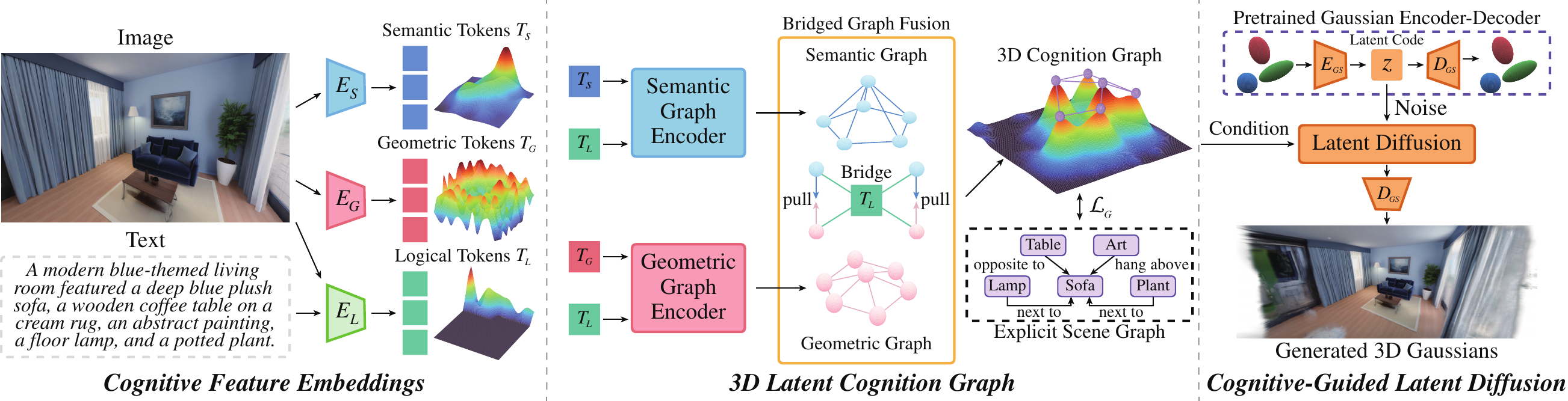}
		\caption{\textbf{Overview of Cog2Gen3D.} The model first extracts multiple cognitive tokens ($T_S, T_G, T_L$). These tokens are structured into a 3D Cognition Graph, where the logical tokens act as a bridge for semantic-geometric alignment. Finally, the awakened 3D cognition steers a latent diffusion process to generate 3D Gaussians.}
		\label{fig2}
	\end{figure}

	\subsection{Semantics-Guided 3D Generation}
	The success of 2D diffusion models has significantly advanced 3D generation. Current methods predominantly leverage their rich semantic priors to supervise 3D representations \cite{poole2022dreamfusion,wang2023score,chen2023fantasia3d,wang2023prolificdreamer,lin2023magic3d,yi2024gaussiandreamer}. For instance, DreamFusion \cite{poole2022dreamfusion} lifts 2D image priors via Score Distillation Sampling (SDS) to iteratively optimize a 3D model. While capable of synthesizing visually impressive objects, these approaches struggle with severe geometric collapse. This limitation arises because they treat 3D generation as multi-view 2D in-painting, lacking an intrinsic perception of physical spatial structures. Therefore, our goal is to introduce geometric cues into the generative process to ensure both appearance fidelity and structural rationality.
	
	\subsection{2D Geometry-Guided 3D Generation}
	To mitigate structural inconsistencies, recent approaches explicitly incorporate 2D geometric priors, such as scene graphs \cite{wang2019planit,dhamo2021graph,zhai2023commonscenes,lin2024instructscene} and layouts \cite{li2019grains,bahmani2023cc3d,po2024compositional,yang2024scenecraft}, to organize global scene composition. For instance, EchoScene \cite{zhai2024echoscene} utilizes topological structures to constrain the semantic interactions among entities, whereas Layout2Scene \cite{chen2025layout2scene} relies on bounding boxes to restrict the relative positions of generated objects. While these structured representations successfully provide a reasonable blueprint to produce roughly arranged scenes, they persistently struggle to ensure precise physical plausibility and strict structural fidelity in complex environments. This limitation fundamentally arises because these methods predominantly model 2D relative spatial relationships rather than the precise absolute metric geometry of the 3D physical world. Consequently, our work focuses on capturing 3D absolute geometry to sculpt structurally rigorous and physically plausible 3D scenes.
	
	\subsection{Spatial Geometric Perception}
	Spatial geometric perception is crucial for 3D generation, as it bridges image-text prompts and structurally rigorous 3D scenes. The core of this perception lies in accurately capturing absolute metric geometry rather than merely coarse topological relations. Recent spatial representation models, such as VGGT \cite{wang2025vggt}, have demonstrated exceptional capabilities in encoding fine-grained 3D spatial geometry. However, effectively integrating such perception into generative models remains a challenge. Existing Multimodal Large Language Models often resort to naive feature concatenation \cite{hong20233d,xu2024pointllm}, which fails to deeply assimilate precise geometric constraints. In contrast, we propose to construct a structured geometric graph and fuse it with a corresponding semantic graph. This design effectively awakens 3D cognition, ensuring both the physical plausibility and structural rationality of the synthesized 3D scenes.
	
	\begin{figure}[t]
		\centering
		\setlength{\abovecaptionskip}{0.1cm} 
		\setlength{\belowcaptionskip}{-0.7cm} 
		\includegraphics[scale=0.145]{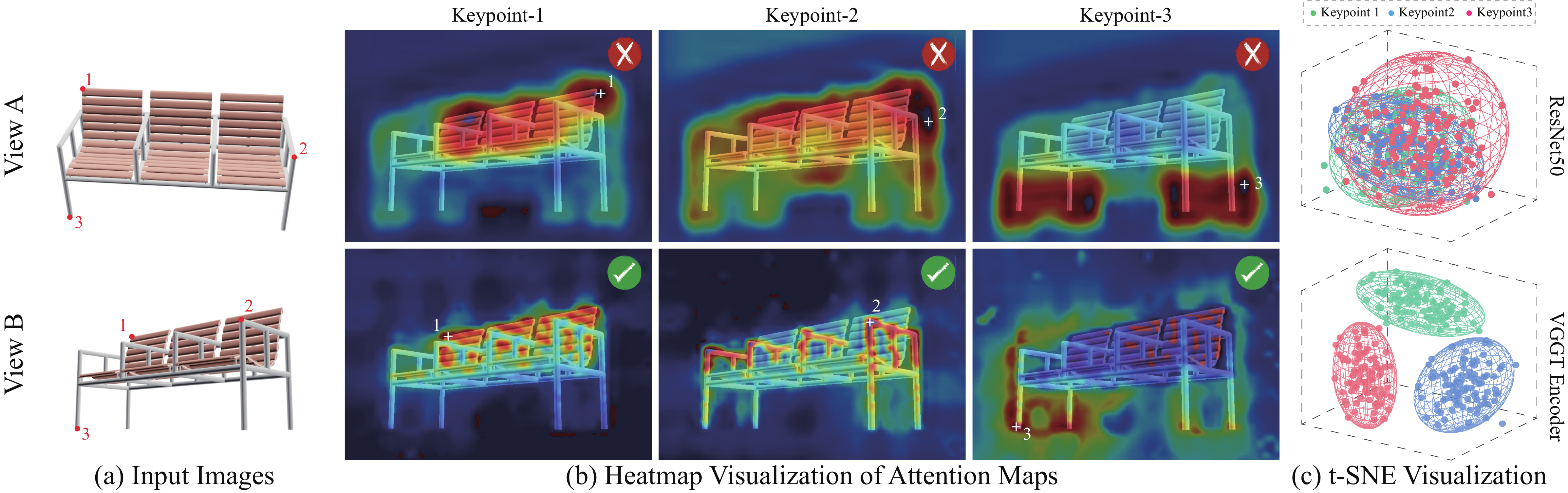}
		\caption{\textbf{Cross-view feature correspondence analysis.} The attention map and the t-SNE visualization demonstrate that the VGGT encoder has superior cross-view geometric consistency. This validates its capability of capturing absolute geometry information, which motivates us to introduce VGGT encoder as our geometric expert.}
		\label{fig3}
	\end{figure}
	
	\section{Our Cog2Gen3D}
	\textbf{Overview.} As shown in \cref{fig2}, our Cog2Gen3D achieves semantically-coherent and geometrically-plausible 3D generation through the following three stages: 
	
	\noindent
	1) \textbf{Cognitive Feature Embeddings.} This is the cognitive representation stage that we transform input images $I$ and texts $T$ into disentangled semantic, geometric, and logical representations $T_S, T_G, T_L$:
	\begin{equation}\small
		\setlength\abovedisplayskip{2pt}
		\setlength\belowdisplayskip{2pt}
		\begin{aligned}
			T_S,T_G,T_L=E_{S,G,L}(I,T),
		\end{aligned}
	\end{equation}
	which serve as the fundamental building blocks for subsequent cognitive reasoning.
	
	\noindent
	2) \textbf{3D Latent Cognition Graph.} This is the 3D cognition construction stage that we encode the extracted features into dual-stream graphs $G_{sem},G_{geo}$ and fuse them via a common-based fusion with the logical bridge $T_L$:
	\begin{equation}\small
		\setlength\abovedisplayskip{4pt}
		\setlength\belowdisplayskip{4pt}
		\begin{aligned}
			G_{sem}=GE(T_S,T_L), G_{geo}=GE(T_G,T_L), G_{cog}=ComFusion(G_{sem},G_{geo},T_L).
		\end{aligned}
	\end{equation}
	The resulting 3D cognition graph $G_{cog}$ precisely captures both the extrinsic semantics and intrinsic geometry of the scene.
	
	\noindent
	3) \textbf{Cognition-Guided Latent Diffusion.} This is the cognition-guided generation stage that the 3D cognition graph acts as a structural condition to steer the latent diffusion process, ultimately generating 3D Gaussians $\hat{\mathcal{G}}$ with high-fidelity appearance and rational structure:
	\begin{equation}\small
		\setlength\abovedisplayskip{4pt}
		\setlength\belowdisplayskip{4pt}
		\begin{aligned}
			\hat{\textbf{z}}_0 = \text{LDM}(\textbf{z}_T \mid G_{cog}), \hat{\mathcal{G}} = D_{GS}(\hat{\textbf{z}}_0).
		\end{aligned}
	\end{equation}
	
	\noindent
	\textbf{Remarks.} The output $\hat{\mathcal{G}}$ denotes the generated 3D Gaussians representing the 3D scene with high-fidelity appearance and rational structure. Our framework utilizes a 3D latent cognition graph as structural conditions to enable semantically-coherent and geometrically-plausible 3D generation in diffusion models.
	
	\subsection{Cognitive Feature Embeddings}
	While conventional generative models rely predominantly on semantic priors, we establish a comprehensive foundation for 3D cognition by integrating semantic and geometric information alongside logical constraints. 
	
	\vspace{1mm}
	\noindent
	\textbf{Semantic Encoder($E_S$).} To capture rich visual appearance, we utilize a pre-trained ResNet50 \cite{he2016deep} as the semantic encoder $E_S$. It extracts high-level visual features from input images and projects them into Semantic Tokens $T_S$, ensuring high-fidelity appearance in the generated 3D Gaussians.
	
	\begin{figure}[t]
		\centering
		\setlength{\abovecaptionskip}{0.1cm} 
		\setlength{\belowcaptionskip}{-0.7cm} 
		\includegraphics[scale=0.45]{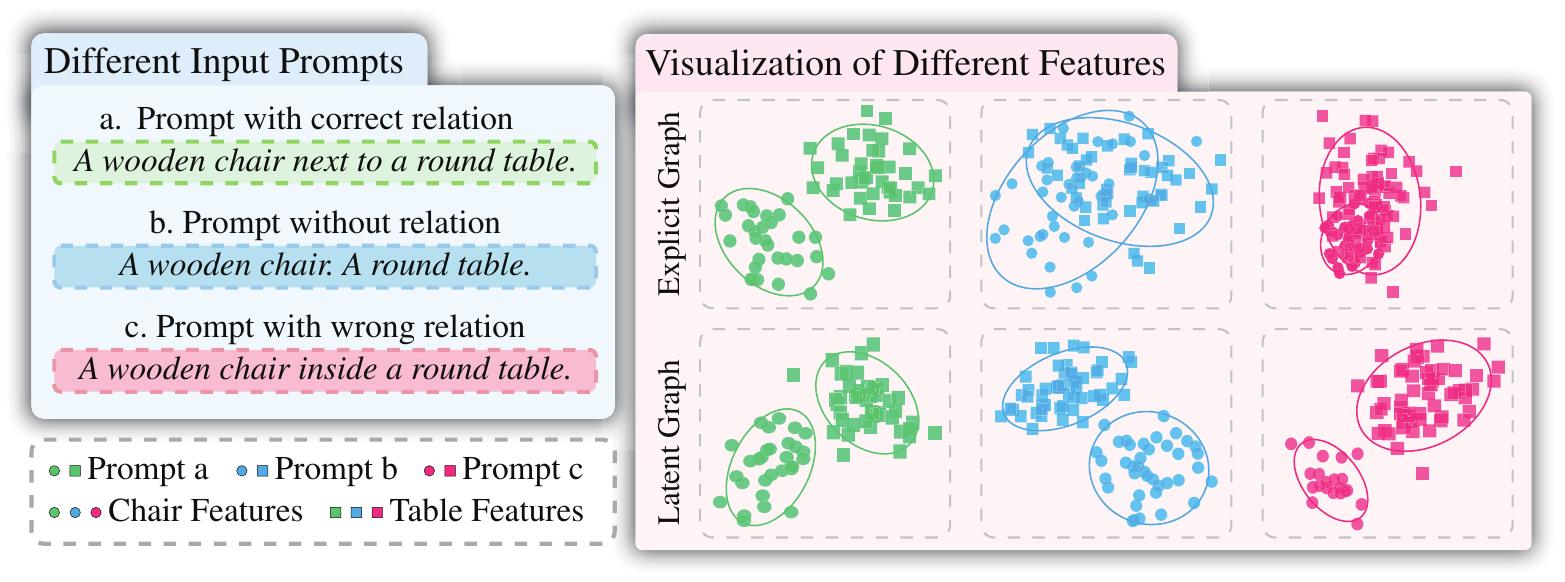}
		\caption{\textbf{Feature distribution of explicit vs. latent scene graphs under prompt perturbations.} Explicit graphs diverge significantly when given missing or incorrect relations compared to the correct baseline. Conversely, our latent scene graph maintains a stable distribution, demonstrating superior robustness to unpromising prompts.}
		\label{fig4}
	\end{figure}
	
	\vspace{1mm}
	\noindent
	\textbf{Geometric Encoder ($E_G$).} As illustrated in \cref{fig3}, we evaluate the feature representations of ResNet50 \cite{he2016deep} and the VGGT encoder \cite{wang2025vggt} by computing cross-view attention between the keypoint-specific features of View A and the spatial feature maps of View B. Visualizations reveal that ResNet50 struggles with feature drift while VGGT distinctly separates keypoint features, demonstrating robust cross-view geometric consistency. Motivated by this capability, we adopt VGGT as our geometric encoder $E_G$, which provides essential geometric grounding, ensuring rigorous structural plausibility and metric accuracy.
	
	\vspace{1mm}
	\noindent
	\textbf{Logical Encoder ($E_L$).} To bridge raw features and structural reasoning, we adopt CLIP ViT and CLIP Text encoders \cite{radford2021learning} as the logical encoder $E_L$. By processing image-text pairs, $E_L$ extracts Logical Tokens $T_L$ that encapsulate high-level relational contexts and abstract concepts, serving as the essential guidance for constructing the 3D cognition of the scene.\newline
	
	Through this tri-stream architecture, we obtain a comprehensive set of cognitive tokens $\{T_S, T_G, T_L\}$, which collectively define the cognitive feature space required for the subsequent 3D Latent Cognition Graph construction.
	
	\subsection{3D Latent Cognition Graph}
	3D generation requires robust guidance equipped with both rich semantics and absolute geometry. \cref{fig4} shows that explicit scene graphs are highly sensitive to noisy inputs, suffering from severe feature divergence under prompt perturbations. To address this, we propose the 3D Latent Cognition Graph. By processing cognitive tokens through a dual-stream latent graph encoder and a common-based fusion mechanism, this module implicitly sculptures a noise-resistant and structurally rigorous cognitive representation.
	
	\vspace{1mm}
	\noindent
	\textbf{Dual-Stream Latent Graph Encoder.}
	To capture coherent appearance and precise geometry, we construct two parallel graphs: a semantic graph and a geometric graph. Both share a similar topological encoding paradigm but differ fundamentally in their input tokens and positional embedding strategies.
	
	\vspace{1mm}
	\noindent
	\emph{Semantic Graph Construction.} We formulate the semantic query $Q_S$ by concatenating tokens $T_S$ with a 2D positional embedding $\text{PE}(x_p, y_p)$. Through cross-attention between $Q_S$ and $T_S$, we extract $n$ initial nodes $N_i^S$ ($i \in [0,n]$). To establish relationships, the cross-attention utilizes the shared logical tokens $T_L$ as logical guidance to formulate semantic edges $E_{ij}^S$ between $N_i^S$ and $N_j^S$ ($i,j \in [0,n]$). An MLP then utilizes the message $m_{ij}^S$ from nodes and corresponding edges to update the nodes, thus obtaining the semantic graph $G_{sem}$:
	\begin{equation}\small
		\setlength\abovedisplayskip{4pt}
		\setlength\belowdisplayskip{4pt}
		\begin{aligned}
			m_{ij}^S=\text{MLP}(N_i^S,N_j^S,E_{ij}^S),\quad G_{sem}=N_i^S+\sum_{j}m_{ij}.
		\end{aligned}
	\end{equation}
	
	\vspace{1mm}
	\noindent
	\emph{Geometric Graph Construction.} To move beyond 2D constraints and model absolute 3D metrics, the geometric query $Q_G$ employs a specialized 3D positional embedding $\text{PE}(x_q, y_q, z_q)$. Here, $z_q$ is a learnable embedding explicitly introduced to capture underlying spatial geometry. Similar to the semantic stream, cross-attention modules extract initial nodes $N_i^G$ and utilize $T_L$ to formulate geometric edges $E_{ij}^G$. An MLP then updates them to produce the geometric graph $G_{geo}$. Driven by the learnable $z_q$ dimension, this graph effectively models absolute 3D metric relationships between entities.
	
	\begin{figure}[t]
		\centering
		\setlength{\abovecaptionskip}{0.1cm} 
		\setlength{\belowcaptionskip}{-0.7cm} 
		\includegraphics[scale=0.368]{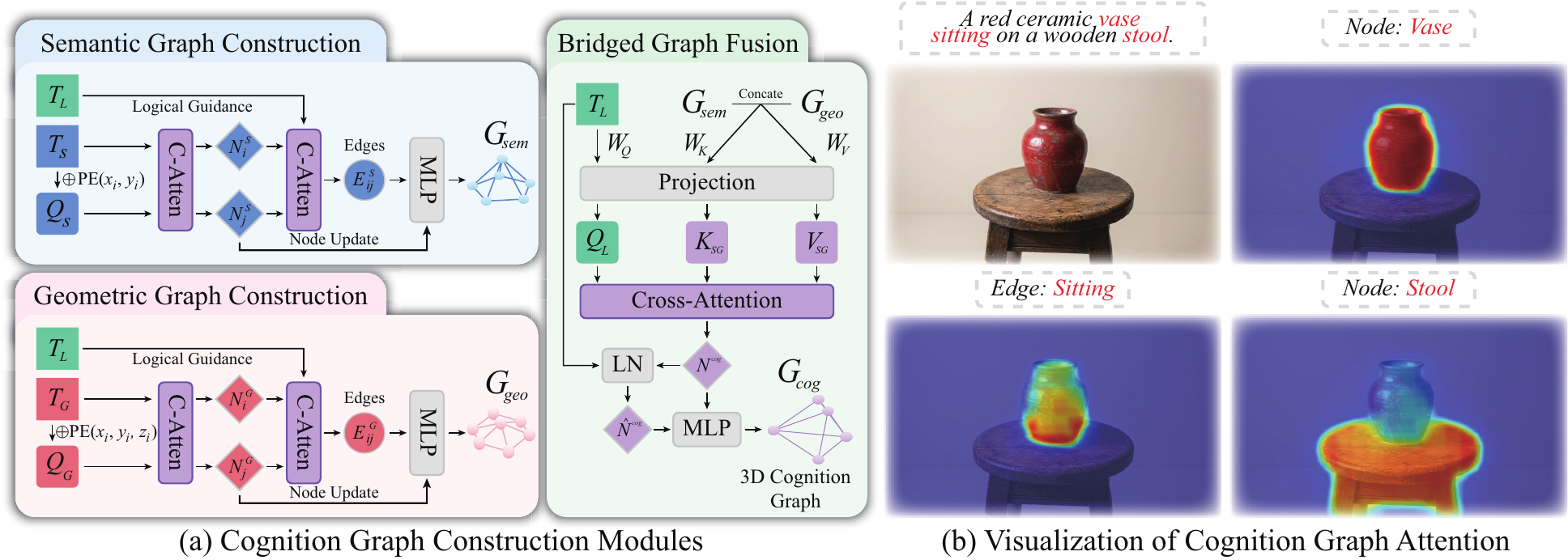}
		\caption{\textbf{Architectural details and interpretability of the 3D Latent Cognition Graph.} (a) The pipeline for constructing cognition graph from cognitive tokens. (b) The correspondence visualization showing how graph components precisely align with 3D entities and spatial boundaries.}
		\label{fig5}
	\end{figure}
	
	\vspace{1mm}
	\noindent
	\textbf{Common-based Cross-Attention Fusion.}
	While the semantic and geometric graphs encapsulate coherent appearance and precise geometry respectively, they remain in separate feature spaces. Since both graphs utilize the same logical tokens $T_L$ for relation edge $E_{ij}$ formulation, their topologies inherently share a common logical foundation.
	
	Exploiting this shared foundation, we introduce a common-based cross-attention fusion mechanism. To integrate the distinct feature spaces, we treat the logical tokens $T_L$ as a unifying anchor. We project $T_L$ to form a shared logical query $\mathbf{Q}_L$, while the semantic and geometric nodes are concatenated along the feature dimension to jointly formulate the keys $\mathbf{K}_{sg}$ and values $\mathbf{V}_{sg}$:
	\begin{equation}\small
		\setlength\abovedisplayskip{4pt}
		\setlength\belowdisplayskip{4pt}
		\begin{aligned}
			\mathbf{Q}_L = T_L \mathbf{W}_Q,\quad \mathbf{K}_{SG} = [G_{sem} || G_{geo}] \mathbf{W}_K,\quad \mathbf{V}_{SG} = [G_{sem} || G_{geo}] \mathbf{W}_V,
		\end{aligned}
	\end{equation}
	where $[\cdot || \cdot]$ denotes the concatenation operation, and $\mathbf{W}_Q, \mathbf{W}_K, \mathbf{W}_V$ are learnable linear projection matrices. The unified 3D cognition graph nodes, denoted as $\mathbf{N}^{cog}$, are then computed via the scaled dot-product cross-attention:
	\begin{equation}\small
		\setlength\abovedisplayskip{4pt}
		\setlength\belowdisplayskip{4pt}
		\begin{aligned}
			\mathbf{N}^{cog} = \text{Softmax}\left( \frac{\mathbf{Q}_L \mathbf{K}_{SG}^T}{\sqrt{d_k}} \right) \mathbf{V}_{SG},
		\end{aligned}
	\end{equation}
	where $d_k$ is the scaling factor representing the channel dimension. In this formulation, the logical query $\mathbf{Q}_L$ acts as an intelligent bridge, adaptively assigning attention weights to extract and align corresponding semantic textures and structural constraints.Finally, we apply a residual connection and a MLP to stabilize the fused representation:
	\begin{equation}\small
		\setlength\abovedisplayskip{4pt}
		\setlength\belowdisplayskip{4pt}
		\begin{aligned}
			\mathbf{\hat{N}}^{cog} = \text{LayerNorm}(T_L + \mathbf{N}^{cog}),\quad \mathbf{G}_{cog} = \text{MLP}(\mathbf{\hat{N}}^{cog}) + \mathbf{\hat{N}}^{cog}.
		\end{aligned}
	\end{equation}
	
	Ultimately, this explicit mathematical fusion yields a holistic 3D cognition graph $\mathbf{G}_{cog}$ that perfectly balances semantic coherence with geometric rationality, ready to structurally guide the subsequent 3D generative process.
	
	\subsection{Cognition-Guided Latent Diffusion}
	With the comprehensive 3D cognition graph formulated, the final stage of our framework is to effectively generate the 3D scenes with semantic fidelity and geometric plausibility. Therefore, we propose a cognition-guided latent diffusion process that operates within a compressed space of 3D Gaussians.
	
	\vspace{1mm}
	\noindent
	\textbf{Conditioned Diffusion Process.}
	Standard diffusion models acting directly on explicit 3D representations are often computationally prohibitive. Therefore, we perform the generative process in a learned latent space. Let the ground-truth latent representation of the 3D scene be denoted as $\mathbf{z}_0$. During the forward diffusion process, Gaussian noise $\epsilon$ is progressively added to $\mathbf{z}_0$ over $t$ timesteps, producing a noisy latent $\mathbf{z}_t$.In the reverse denoising process, our goal is to predict and remove this noise. Crucially, instead of relying on rudimentary text or layout conditions, we inject our fused cognition graph, $\mathbf{G}_{cog}$, as the structural condition. The denoising network predicts the added noise as:
	\begin{equation}\small
		\setlength\abovedisplayskip{4pt}
		\setlength\belowdisplayskip{4pt}
		\begin{aligned}
			\hat{\epsilon} = \epsilon_\theta(\mathbf{z}_t, t, \mathbf{G}_{cog}).
		\end{aligned}
	\end{equation}
	
	Within the denoising network, the high-fidelity semantics and plausible geometry embedded in $\mathbf{G}_{cog}$ effectively guide the generation of the latent space. This cognition-guided paradigm effectively mitigates the geometric ambiguity and layout distortion commonly observed in standard 2D-prior-based generation.
	
	\vspace{1mm}
	\noindent
	\textbf{Latent-Gaussian Encoder-Decoder.}
	To bridge the gap between the compact latent space and the explicit 3D representation, we employ a pre-trained latent-Gaussian encoder-decoder architecture as shown in \cref{fig2}. A standard 3D Gaussian splatting scene is explicitly parameterized by a set of Gaussians $\mathcal{G}$. The Gaussian encoder $E_{GS}$ compresses the high-dimensional Gaussians into the compact latent space: $\mathbf{z}_0 = E_{GS}(\mathcal{G})$. Then we apply noise to obtain $\mathbf{z}_t$. Conversely, after the conditioned diffusion process yields the denoised latent $\hat{\mathbf{z}}_0$, the decoder $D_{GS}$ projects it back into the explicit 3D Gaussian space: $\hat{\mathcal{G}}=D_{GS}(\hat{\mathbf{z}}_0)$. This symmetric design ensures that our model can efficiently learn complex spatial distributions while ultimately generating high-fidelity 3D Gaussians.
	
	\subsection{Optimization}
	Our proposed Cog2Gen3D framework is supervised by training loss consists of three primary components: a latent diffusion loss $\mathcal{L}_{diff}$, an explicit node grounding loss $\mathcal{L}_{G}$, and a 3D Gaussian reconstruction loss $\mathcal{L}_{recon}$.
	
	\vspace{1mm}
	\noindent
	\textbf{Latent Diffusion Loss $\mathcal{L}_{diff}$.} To train the conditioned denoising network, we employ the standard latent diffusion objective, minimizing the mean squared error between the added noise $\epsilon$ and the predicted noise $\hat{\epsilon}$:
	\begin{equation}\small
		\setlength\abovedisplayskip{4pt}
		\setlength\belowdisplayskip{4pt}
		\begin{aligned}
			\mathcal{L}_{diff} = \mathbb{E}_{\mathbf{z}_0, \epsilon, t} \left[ || \epsilon - \hat{\epsilon} ||_2^2 \right].
		\end{aligned}
	\end{equation}
	
	\vspace{1mm}
	\noindent
	\textbf{Explicit Node Grounding Loss $\mathcal{L}_{g}$.} Explicit scene graphs typically lack precise spatial grounding and contain sparse relations. Supervising latent edges with such limited data would bottleneck spatial reasoning. Therefore, we discard edge supervision and solely anchor the semantic identities of the nodes. Crucially, the dense nodes in our cognition graph do not strictly align with the sparse entities in the explicit graph. To address this, we dynamically select the top-$K$ most critical latent nodes via cross-attention weight ranking that correspond to the $K$ explicit entities. We pass these selected nodes through a classification head to output semantic probabilities $p_i^{sem}$. The loss is computed as a Cross-Entropy (CE) objective against the explicit semantic labels $y_i^{sem}$:
	\begin{equation}\small
		\setlength\abovedisplayskip{4pt}
		\setlength\belowdisplayskip{4pt}
		\begin{aligned}
			\mathcal{L}_g = \frac{1}{K} \sum_{i=1}^{K} \text{CE}(p_i^{sem}, y_i^{sem}).
		\end{aligned}
	\end{equation}
	
	This minimalist top-$K$ supervision ensures semantic fidelity while granting the latent edges absolute freedom to autonomously infer complex 3D topology.
	
	\vspace{1mm}
	\noindent
	\textbf{3D Gaussian Reconstruction Loss $\mathcal{L}_{recon}$.} To guarantee metric precision and visual fidelity, we supervise the Gaussians $\hat{\mathcal{G}}$ using the ground-truth $\mathcal{G}$. This supervision is applied in the image space to ensure multi-view consistency. For a set of viewpoints $v$, we render the predicted Gaussians into images $\hat{I}_v$ and compare them against the GT renders $I_v$, calculating L1 loss and D-SSIM loss:
	\begin{equation}\small
		\setlength\abovedisplayskip{4pt}
		\setlength\belowdisplayskip{4pt}
		\begin{aligned}
			\mathcal{L}_{recon} = \sum_{v} \left( \lambda_{L1} || \hat{I}_v - I_v ||_1 + \lambda_{ssim} (1 - \text{SSIM}(\hat{I}_v, I_v)) \right).
		\end{aligned}
	\end{equation}
	
	\vspace{1mm}
	\noindent
	\textbf{Total Loss.} The entire framework is optimized end-to-end using a weighted sum of the above three losses, which is formulated as:
	\begin{equation}\small
		\setlength\abovedisplayskip{4pt}
		\setlength\belowdisplayskip{4pt}
		\begin{aligned}
			\mathcal{L}_{total} = \lambda_1 \mathcal{L}_{diff} + \lambda_2 \mathcal{L}_g + \lambda_3 \mathcal{L}_{recon},
			\label{eq1}
		\end{aligned}
	\end{equation}
	
	where $\lambda_1, \lambda_2, \lambda_3$ are hyperparameters balancing the contributions of diffusion fidelity, structural cognition, and visual reconstruction.
	
	\section{Training Datasets and Pipeline}
	\subsection{Our CogSG-3D Dataset}
\begin{figure}[t]
		\centering
		\setlength{\abovecaptionskip}{0.1cm} 
		\setlength{\belowcaptionskip}{-0.7cm} 
		\includegraphics[scale=0.2]{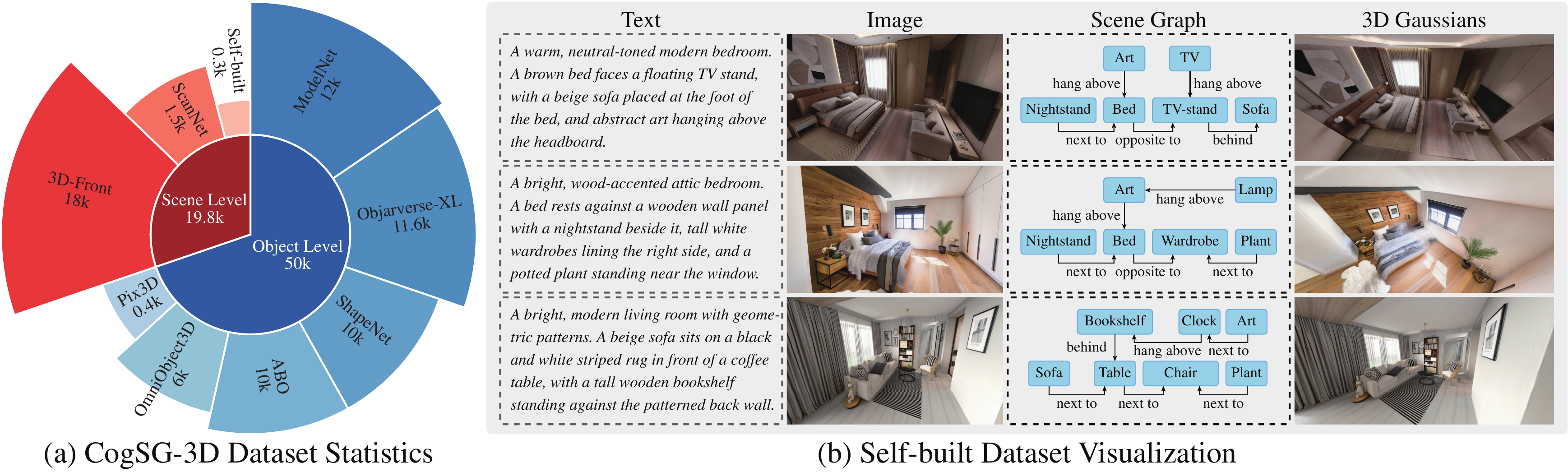}
		\caption{Statistics and examples of our proposed CogSG-3D dataset.}
		\label{fig6}
	\end{figure}

	As shown in \cref{fig6}, to effectively train our Cog2Gen3D, we propose CogSG-3D, a comprehensive dataset aggregating pre-eminent public 3D datasets and self-built supplementary data from Marble World Labs \cite{worldlabs2025marble}. It covers two categories: object-level (ShapeNet \cite{chang2015shapenet}, Objaverse-XL \cite{deitke2023objaverse}, OmniObject3D \cite{wu2023omniobject3d}, Pix3D \cite{sun2018pix3d}, ABO \cite{collins2022abo}, ModelNet \cite{wu20153d}) and scene-level (ScanNet \cite{dai2017scannet}, 3D-Front \cite{fu20213d}), ensuring diverse semantic and geometric distributions. To unify these heterogeneous source formats into a standardized format for training, our data processing pipeline encompasses three steps: 1) \emph{Image prompt rendering.} We normalize the canonical poses of all 3D models and render 2D multi-view images to serve as visual prompts. 2) \emph{Text prompt acquisition}. For datasets lacking explicit annotations, we leverage advanced Vision-Language Models (Gemini) to generate dense descriptive texts, ensuring rigorous vision-language alignment. 3) \emph{Unified 3D Gaussians representation.} We convert all 3D ground truths into directly optimizable 3D Gaussians, overcoming the topological constraints of traditional meshes.
	
	\subsection{Training Pipeline}
	We design a progressive, three-stage training paradigm to ensure stable convergence and optimal semantic-geometric alignment:
	
	\vspace{1mm}
	\noindent
	\textbf{Stage 1: Geometric-Latent Alignment.} This stage aims to acquire latent representations of 3D scenes and a robust decoder. We pretrain a Gaussians auto-decoder where the encoder $E_{GS}$ maps 3D Gaussians into a compact latent variable $z$ and the decoder $D_{GS}$ is simultaneously optimized to flawlessly reconstruct the 3D Gaussians from $z$, preserving rich semantic and geometric details.
	
	\vspace{1mm}
	\noindent
	\textbf{Stage 2: Cognitive-Generative Alignment.} The objective here is to synthesize semantically and geometrically coherent 3D latents guided by multimodal priors. Keeping the autoencoder frozen, we exclusively train the Latent Diffusion Model (LDM) in the $z$-space. The denoising network is conditioned on the 3D Latent Cognition Graph to enforce strict structural constraints.
	
	\vspace{1mm}
	\noindent
	\textbf{Stage 3: End-to-End Fine-Tuning.} This stage harmonizes the generative latent space with the explicit 3D rendering space. We unfreeze the decoder $D_{GS}$ and jointly optimize it with the LDM end-to-end. This mitigates feature mismatch between synthesized latents and the decoder's expected input distribution, significantly enhancing the overall visual fidelity and geometric precision.
	
	\section{Experiments}
	\subsection{Implementation Details}
	Our 3D latent cognition graph contains 64 hidden nodes by default, which can be manually adjusted according to scene complexity. For the diffusion process, we employ $T=1000$ training timesteps, while a 50-step DDIM sampler is utilized for efficient inference. The total loss is optimized using the AdamW optimizer with a initial learning rate of $1\times 10^{-5}$, where the loss weights in \cref{eq1} are set as: $\lambda_1=0.8$, $\lambda_2=0.2$, and $\lambda_3=0.8$. All training and testing procedures are conducted on 8 NVIDIA A800 GPUs.
	
	\begin{table}[t]
		\centering
		\setlength{\abovecaptionskip}{0.0cm}
		\setlength{\belowcaptionskip}{0.0cm}
		\captionsetup{font={scriptsize}}
		\addvbuffer[0pt -6pt]{	
			\begin{minipage}[t]{0.512\textwidth}
				\vspace{0pt} 
				\centering
				\caption{Text-to-3D comparison.}
				\label{tab1}
				\setlength{\tabcolsep}{3.0pt}
				\renewcommand{\arraystretch}{0.9}
				\resizebox{\linewidth}{!}{
					\begin{tabular}{ccccc}
						\toprule
						\multirow{2}{*}{Method} & \multicolumn{4}{c}{T3Bench \cite{he2023t}} \\
						\cmidrule(lr){2-5}
						& Single Obj.$\uparrow$ & Single w/ Surr.$\uparrow$ & Multi Obj.$\uparrow$ & Average$\uparrow$ \\
						\midrule
						DreamFusion\cite{poole2022dreamfusion} & 24.4 & 19.8 & 11.7 & 18.7 \\
						Magic3D\cite{lin2023magic3d} & 37 & 35.4 & 25.7 & 32.7 \\
						SJC\cite{wang2023score} & 24.7 & 19.8 & 11.7 & 18.7 \\
						Fantasia3D\cite{chen2023fantasia3d} & 26.4 & 27 & 18.5 & 24 \\
						ProlificDreamer\cite{wang2023prolificdreamer} & 49.4 & 44.8 & 35.8 & 43.3 \\
						GaussianDreamer\cite{yi2024gaussiandreamer} & 54 & 48.6 & 34.5 & 45.7 \\
						\textbf{Ours} & \textbf{58.3} & \textbf{57.9} & \textbf{53.6} & \textbf{56.6} \\
						\bottomrule
					\end{tabular}
				}
			\end{minipage}\hfill 
			\begin{minipage}[t]{0.48\textwidth}
				\vspace{0pt} 
				\centering
				\caption{Image-to-3D objects comparison.}
				\label{tab2}
				\setlength{\tabcolsep}{2.5pt}
				\renewcommand{\arraystretch}{0.9}
				\resizebox{\linewidth}{!}{
					\begin{tabular}{ccccccc}
						\toprule
						\multirow{2}{*}{Method} & \multicolumn{3}{c}{ShapeNet \cite{chang2015shapenet}} & \multicolumn{3}{c}{OmniObject3D \cite{wu2023omniobject3d}} \\
						\cmidrule(lr){2-4} \cmidrule(lr){5-7}
						& FID$\downarrow$ & KID(\%)$\downarrow$ & MMD(\%)$\downarrow$ & FID$\downarrow$ & KID(\%)$\downarrow$ & MMD(\textperthousand)$\downarrow$ \\
						\midrule
						EG3D\cite{chan2022efficient} & 28.12 & 1.25 & 7.14 & 41.56 & 1.13 & 28.21 \\
						GET3D\cite{gao2022get3d} & 38.62 & 1.85 & 6.45 & 49.41 & 1.53 & 13.57 \\
						DiffRF\cite{muller2023diffrf} & 98.53 & 6.14 & 8.59 & 147.59 & 8.82 & 16.07 \\
						DiffTF\cite{cao2023large} & 26.58 & 1.15 & 6.15 & 25.36 & 0.81 & 6.64 \\
						DiffGS\cite{zhou2024diffgs} & 23.45 & 1.05 & 5.31 & 27.35 & 0.84 & 7.05 \\
						LN3Diff\cite{lan2024ln3diff} & 21.51 & 1.08 & 5.25 & 21.42 & 0.73 & 4.87 \\
						\textbf{Ours} & \textbf{14.54} & \textbf{0.67} & \textbf{3.25} & \textbf{15.94} & \textbf{0.58} & \textbf{3.02} \\
						\bottomrule
					\end{tabular}
				}
			\end{minipage}
		}
	\end{table}
	
	\begin{figure}[t]
		\centering
		\setlength{\abovecaptionskip}{0.1cm} 
		\setlength{\belowcaptionskip}{-0.7cm} 
		\includegraphics[scale=0.095]{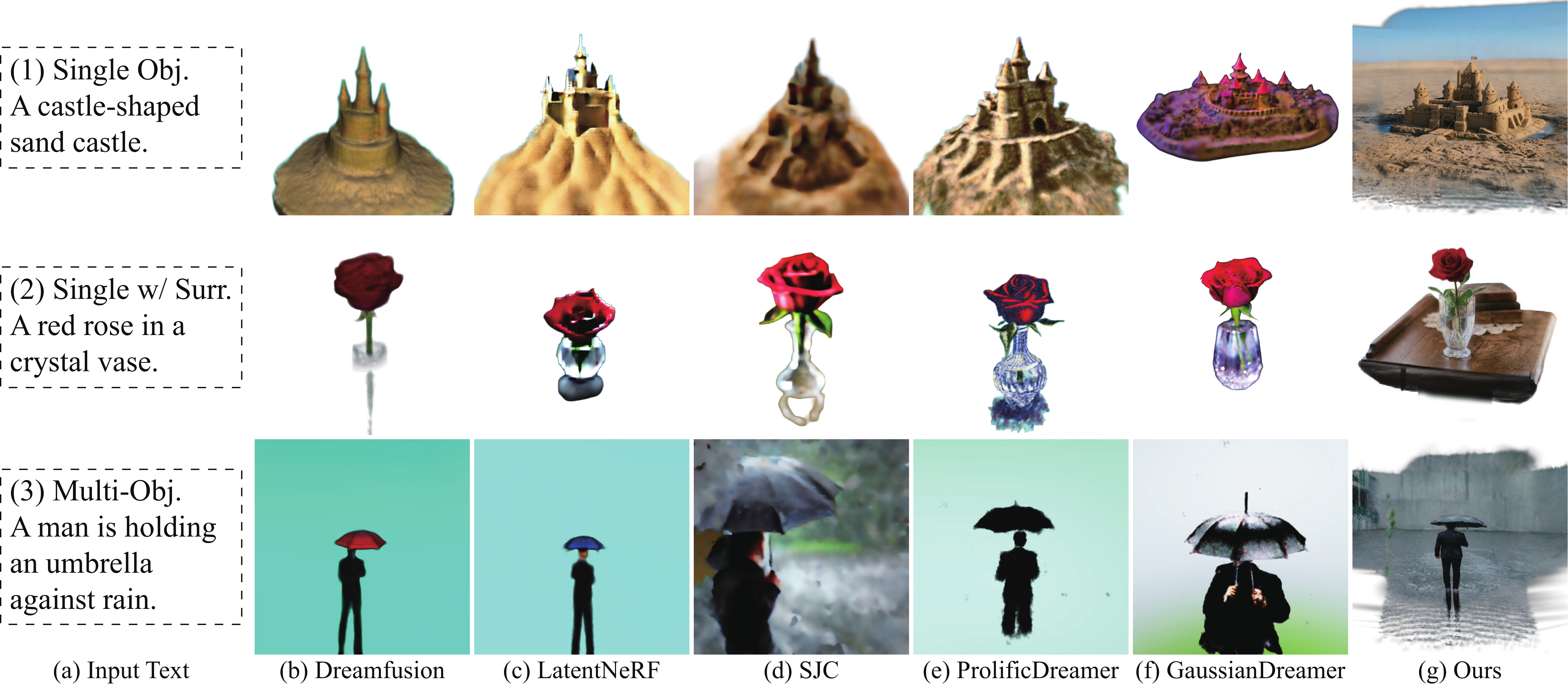}
		\caption{Visualization of Text-to-3D on the T3-Bench dataset.}
		\label{fig7}
	\end{figure}
	
	\subsection{Comparison Experiments}
	
	\vspace{1mm}
	\noindent
	\textbf{Text-to-3D Generation.}
	We evaluate text-to-3D generation on T3Bench \cite{he2023t} against SOTA methods \cite{poole2022dreamfusion,lin2023magic3d,wang2023score,chen2023fantasia3d,wang2023prolificdreamer,yi2024gaussiandreamer}. We adopt the T3Bench quality scores across three complexity levels and their average. As shown in \cref{tab1} and \cref{fig6}, the results demonstrate the superiority of our approach. Quantitatively, Cog2Gen3D achieves the highest scores across all metrics, with notable margins in the challenging multi-object task. Qualitatively, while baseline methods often suffer from blurred details, geometric distortions, or structural collapse in complex scenes due to the lack of spatial constraints, our method consistently produces high-fidelity 3D assets. Benefiting from the structural constraints of our 3D cognition graph, Cog2Gen3D maintains precise geometries and coherent multi-entity relationships, demonstrating significant advantages in holistic 3D scene generation.
	
	\begin{figure}[t]
		\centering
		\setlength{\abovecaptionskip}{0.1cm} 
		\setlength{\belowcaptionskip}{-0.7cm} 
		\includegraphics[scale=0.127]{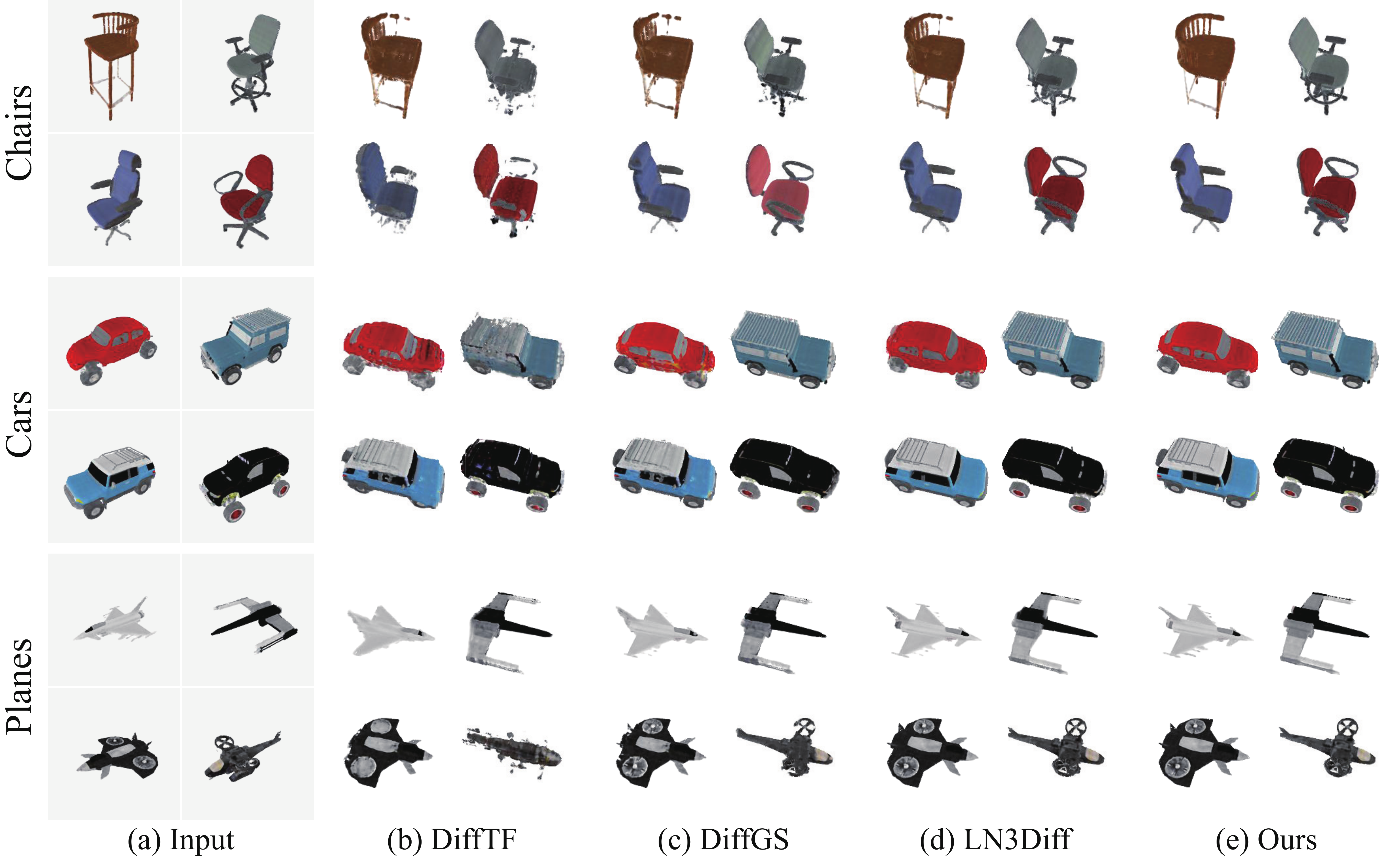}
		\caption{Visualization of Image-to-3D objects on the ShapeNet dataset.}
		\label{fig8}
	\end{figure}
	
	\vspace{1mm}
	\noindent
	\textbf{Image-to-3D Object Generation.}
	We evaluate image-to-3D object generation on ShapeNet \cite{chang2015shapenet} and OmniObject3D \cite{wu2023omniobject3d} against leading baselines \cite{chan2022efficient,gao2022get3d,muller2023diffrf,cao2023large,zhou2024diffgs,lan2024ln3diff}. We adopt standard evaluation metrics such as FID, KID, and MMD. \cref{tab2} and \cref{fig8} shows that Cog2Gen3D establishes a clear advantage over existing baselines. Quantitatively, our model achieves the top scores across all metrics and datasets. Qualitatively, our framework reliably reconstructs detailed 3D assets. Leveraging the semantic and geometric perception of our 3D cognition graph, Cog2Gen3D excels at preserving high-fidelity appearances and plausible geometric structures.
	
	\vspace{1mm}
	\noindent
	\textbf{Image-to-3D Scene Generation.}
	We validate complex scene generation on 3D-Front \cite{fu20213d} and our CogSG-3D datasets against semantic \cite{chung2023luciddreamer,chen2024single,ardelean2025gen3dsr,han2025reparo,huang2025midi} and 2D geometry-guided \cite{zhai2024echoscene,chen2025layout2scene} baselines. We employ standard metrics such as Chamfer Distance (CD), F-Score, and IoU to evaluate the structural plausibility and visual fidelity of the generated 3D scenes. The quantitative results in \cref{tab3} shows that Cog2Gen3D consistently outperforms all competing baselines. Visual comparisons in \cref{fig9} reveal that existing approaches often struggle with scale inconsistencies, chaotic spatial layouts, or structural collapse when synthesizing cluttered spaces due to the lack of spatial awareness. Conversely, our model generates highly realistic and well-organized 3D environments by embedding semantic and absolute geometric cognitive features into the generation process.

	\subsection{Ablation Study}
	\vspace{1mm}
	\noindent
	\textbf{Effectiveness of Cognitive Feature Embeddings.}
	A user study (\cref{tab4}) evaluates the distinct roles of the three cognitive tokens across Semantic Fidelity (SemF), Geometric Plausibility (GeoP), and Relational Coherence (RelC). Ablating the semantic, geometric, or logical tokens severely degrades SemF, GeoP, or RelC, respectively. This direct correspondence confirms that integrating all three embeddings is essential for synthesizing scenes with high-fidelity semantics, plausible geometry, and coherent logic.

	\vspace{1mm}
	\noindent
	\textbf{Superiority of Latent Graph Construction.}
	We validate our structured topology by replacing it with a flattened token sequence (\cref{tab5}). This ablation degrades performance, indicating that flat sequences fail to capture complex 3D spatial dependencies. Conversely, our structured graph effectively models semantic-geometric interactions, significantly enhancing the structural rationality and geometric plausibility of generated scenes.
	
	\vspace{1mm}
	\noindent
	\textbf{Impact of Dual-Stream Scene Graphs.}
	We ablate the semantic and geometric streams to assess their contributions. \cref{fig10} shows that removing either stream degrades performance. Moreover, omitting semantic graphs compromises texture fidelity and lacking geometric graphs causes severe structural distortions. This confirms both streams are indispensable for high-quality 3D generation.
	
	\begin{table}[t]
		\centering
		\setlength{\abovecaptionskip}{0.0cm}
		\setlength{\belowcaptionskip}{0.0cm}
		\captionsetup{font={scriptsize}}
		\addvbuffer[0pt -6pt]{	
			\begin{minipage}[t]{0.487\textwidth}
				\vspace{0pt} 
				\centering
				\caption{Image-to-3D scenes comparison.}
				\label{tab3}
				\setlength{\tabcolsep}{3.0pt}
				\renewcommand{\arraystretch}{0.8}
				\resizebox{\linewidth}{!}{
					\begin{tabular}{cccc}
						\toprule
						\multirow{2}{*}{Method} & \multicolumn{3}{c}{3D-Front \cite{fu20213d}} \\
						\cmidrule(lr){2-4}
						& Chamfer Dist.$\downarrow$ & F-Score$\uparrow$ & IoU$\uparrow$ \\
						\midrule
						LucidDreamer\cite{chung2023luciddreamer} & 0.083 & 50.79 & 0.536 \\
						SSR\cite{chen2024single}      & 0.140 & 39.76 & 0.311 \\
						Gen3DSR\cite{ardelean2025gen3dsr}  & 0.123 & 40.07 & 0.363 \\
						REPARO\cite{han2025reparo}   & 0.129 & 41.68 & 0.339 \\
						MIDI\cite{huang2025midi}     & 0.080 & 50.19 & 0.518 \\
						EchoScene\cite{zhai2024echoscene}  & 0.105 & 45.62 & 0.458 \\
						Layout2Scene\cite{chen2025layout2scene}   & 0.094 & 48.36 & 0.492 \\
						\textbf{Ours} & \textbf{0.063} & \textbf{58.43} & \textbf{0.682} \\
						\bottomrule
					\end{tabular}
				}
			\end{minipage}\hfill 
			\begin{minipage}[t]{0.5\textwidth}
				\vspace{0pt} 
				\centering
				\caption{Ablation study on cognitive tokens.}
				\label{tab4}
				\setlength{\tabcolsep}{3.0pt}
				\renewcommand{\arraystretch}{0.8}
				\resizebox{\linewidth}{!}{
					\begin{tabular}{cccccc}
						\toprule
						\multicolumn{3}{c}{Cognitive Tokens} & \multicolumn{3}{c}{User Study $\uparrow$} \\
						\cmidrule(lr){1-3} \cmidrule(lr){4-6}
						Semantic & Geometric & Logical & SemF & GeoP & RelC \\
						\midrule
						\checkmark &            &            & 4.12 & 2.38 & 2.15 \\
						& \checkmark &            & 2.23 & 4.08 & 2.74 \\
						&            & \checkmark & 2.41 & 2.58 & 3.92 \\
						\checkmark & \checkmark &            & 4.38 & 4.35 & 2.82 \\
						\checkmark &            & \checkmark & 4.42 & 2.75 & 4.28 \\
						& \checkmark & \checkmark & 3.17 & 4.31 & 4.34 \\
						\midrule
						\checkmark & \checkmark & \checkmark & \textbf{4.65} & \textbf{4.58} & \textbf{4.62} \\
						\bottomrule
					\end{tabular}
				}
			\end{minipage}
		}
	\end{table}
	
	\begin{figure}[t]
		\centering
		\setlength{\abovecaptionskip}{0.1cm} 
		\setlength{\belowcaptionskip}{-0.7cm} 
		\includegraphics[scale=0.045]{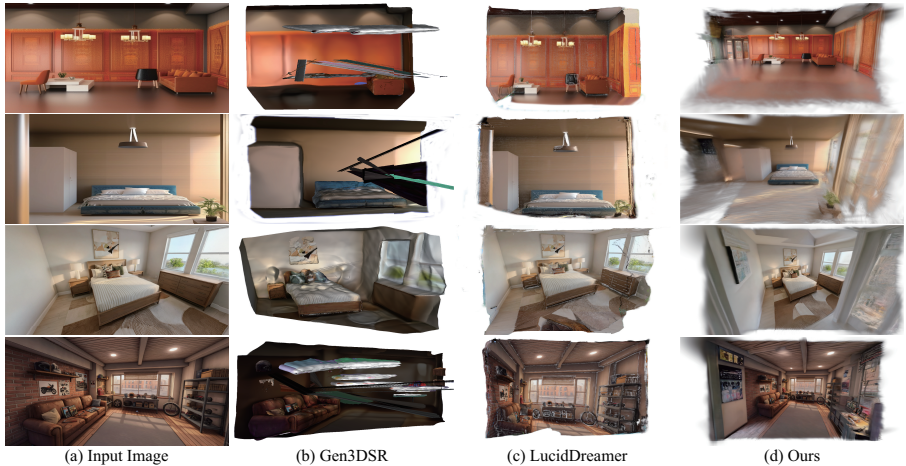}
		\caption{Visualization of Image-to-3D Scenes on 3D-Front and our CogSG-3D datasets.}
		\label{fig9}
	\end{figure}
	
	\subsection{Discussion}
	\vspace{1mm}
	\noindent
	\textbf{Correspondence between the Cognition Graph and 3D Scenes.}
	To interpret the latent cognition graph, we compute attention maps between the generated 3D scenes and node/edge embeddings. \cref{fig5} shows that high-attention regions precisely align with target objects and spatial boundaries. This confirms that the graph acts as a structured cognitive map, successfully localizing semantic-geometric entities to bridge abstract prompts with 3D scenes.

	\vspace{1mm}
	\noindent
	\textbf{Influence of the Geometry Perception Backbone.}
	We evaluate different geometry encoders in \cref{tab6}. The VGGT encoder \cite{wang2025vggt} achieves optimal performance by excelling at extracting robust geometric features and spatial structures. This provides precise absolute geometry guidance for the diffusion process, significantly enhancing the geometric plausibility of the generated 3D scenes.
	
	\vspace{1mm}
	\noindent
	\textbf{Effectiveness of Common-based Fusion Strategy.}
	We compare our common-based fusion against standard concatenation and weighted fusion. \cref{tab7} shows that our approach effectively synergizes semantic and geometric features. This deep alignment preserves modality-specific priors, yielding the highest generation quality in both semantic fidelity and geometric plausibility.
	
	\begin{table}[t]
		\centering
		\setlength{\abovecaptionskip}{0.05cm}
		\setlength{\belowcaptionskip}{0.0cm}
		\captionsetup{font={scriptsize}}
		\addvbuffer[0pt -10pt]{	
			\begin{minipage}[t]{0.349\textwidth}
				\vspace{0pt} 
				\centering
				\caption{Graph cons. ablation.}
				\label{tab5}
				\setlength{\tabcolsep}{2.0pt}
				\renewcommand{\arraystretch}{1.0}
				\resizebox{\linewidth}{!}{
					\begin{tabular}{cccc}
						\toprule
						\multirow{2}{*}{Settings} & \multicolumn{3}{c}{3D-Front \cite{fu20213d}} \\
						\cmidrule(lr){2-4}
						& CD $\downarrow$ & F-Score $\uparrow$ & IoU $\uparrow$ \\
						\midrule
						w/o Graph & 0.089 & 49.16 & 0.503 \\
						\textbf{w/. Graph} & \textbf{0.063} & \textbf{58.43} & \textbf{0.683} \\
						\bottomrule
					\end{tabular}
				}
			\end{minipage}\hfill
			\begin{minipage}[t]{0.32\textwidth}
				\vspace{0pt} 
				\centering
				\caption{Geometry encoders.}
				\label{tab6}
				\setlength{\tabcolsep}{2.0pt}
				\renewcommand{\arraystretch}{0.9}
				\resizebox{\linewidth}{!}{
					\begin{tabular}{cccc}
						\toprule
						\multirow{2}{*}{Settings} & \multicolumn{3}{c}{3D-Front \cite{fu20213d}} \\
						\cmidrule(lr){2-4}
						& CD $\downarrow$ & F-Score $\uparrow$ & IoU $\uparrow$ \\
						\midrule
						ResNet50 & 0.091 & 48.75 & 0.498 \\
						CLIP ViT-L & 0.076 & 53.88 & 0.591 \\
						\textbf{VGGT} & \textbf{0.063} & \textbf{58.43} & \textbf{0.683} \\
						\bottomrule
					\end{tabular}
				}
			\end{minipage}\hfill
			\begin{minipage}[t]{0.304\textwidth}
				\vspace{0pt} 
				\centering
				\caption{Fusion strategy.}
				\label{tab7}
				\setlength{\tabcolsep}{2.0pt}
				\renewcommand{\arraystretch}{0.9}
				\resizebox{\linewidth}{!}{
					\begin{tabular}{cccc}
						\toprule
						\multirow{2}{*}{Settings} & \multicolumn{3}{c}{3D-Front \cite{fu20213d}} \\
						\cmidrule(lr){2-4}
						& CD $\downarrow$ & F-Score $\uparrow$ & IoU $\uparrow$ \\
						\midrule
						Concat. & 0.085 & 50.41 & 0.523 \\
						Weighted & 0.072 & 55.12 & 0.635 \\
						\textbf{Common} & \textbf{0.063} & \textbf{58.43} & \textbf{0.683} \\
						\bottomrule
					\end{tabular}
				}
			\end{minipage}
		}
	\end{table}
	
	\begin{figure}[t]
		\centering
		\setlength{\abovecaptionskip}{0.1cm} 
		\setlength{\belowcaptionskip}{-0.7cm} 
		\includegraphics[scale=0.057]{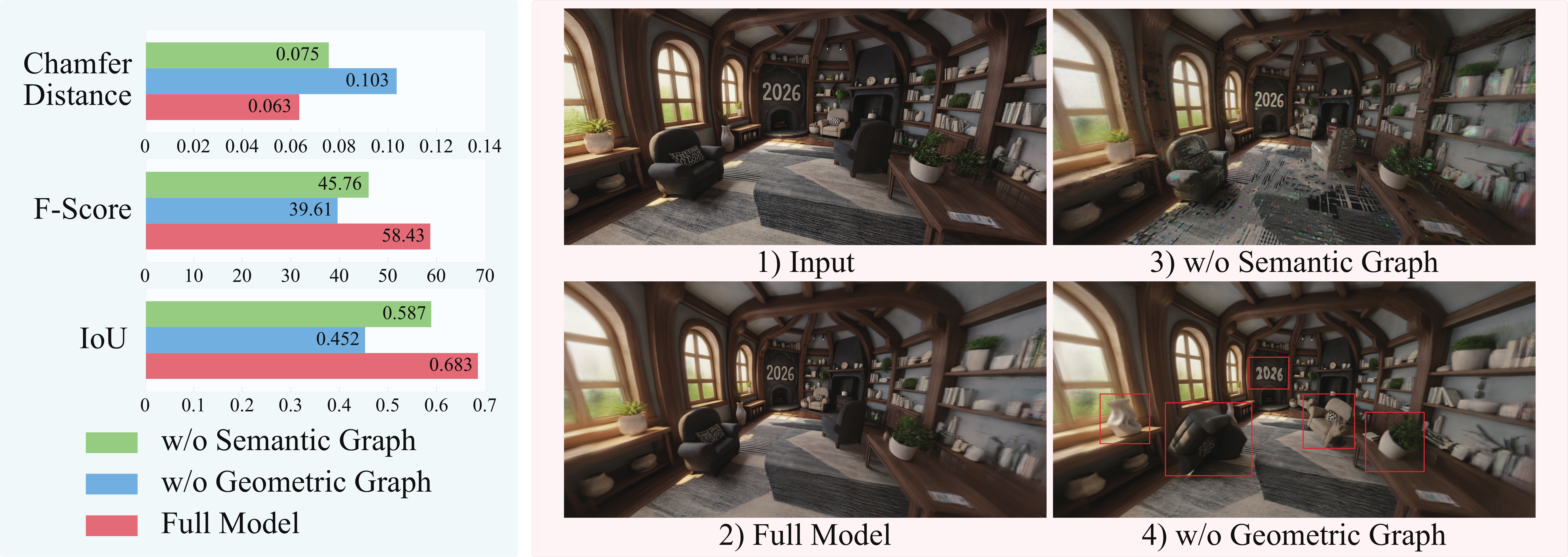}
		\caption{\textbf{Ablation of dual-stream scene graphs.} Quantitative metrics drop noticeably without either stream. Qualitative examples reveal that omitting semantic or geometric graphs causes severe texture degradation or structural collapse, respectively.}
		\label{fig10}
	\end{figure}
	
	\vspace{1mm}
	\noindent
	\textbf{Limitations.}
	While Cog2Gen3D demonstrates strong capabilities in generating high-quality and geometrically plausible static 3D scenes, it currently struggles with dynamic 4D generation. This limitation primarily arises from the absence of temporal modeling within our framework. Our 3D cognition graph and Gaussian representation \cite{kerbl20233d} are restricted to static constraints, failing to capture motion or topological evolution. Future work will integrate spatio-temporal graphs and 4D Gaussian Splatting \cite{wu20244d} to enable the generation of spatially grounded and temporally consistent dynamic scenes.
	
	\section{Conclusion}
	In this work, we proposed Cog2Gen3D, a 3D cognition-guided diffusion framework designed to sculpture semantic-geometric cognition for high-quality 3D generation. By integrating absolute geometric priors and semantic constraints, our method effectively resolves the challenges of scale inconsistency and the absence of spatial awareness prevalent in existing models. We introduced cognitive feature embeddings to map multi-modal information into a unified semantic-geometric space, and proposed the 3D latent cognition graph to capture complex spatial topological relationships and structural dependencies within 3D environments. Furthermore, we constructed the CogSG-3D dataset, providing extensive explicit scene graph and 3D Gaussian annotations to support training. The superiority of our approach has been demonstrated through extensive experiments across multiple tasks, including text-to-3D, image-to-3D object, and complex scene generation, affirming its effectiveness in modeling the physical world.

	%
	%
	\bibliographystyle{splncs04}
	\bibliography{main}

\begin{thebibliography}{10}

\bibitem{zhang2023adding}
Lvmin Zhang, Anyi Rao, and Maneesh Agrawala.
\newblock Adding conditional control to text-to-image diffusion models.
\newblock In {\em Proceedings of the IEEE/CVF international conference on
  computer vision}, pages 3836--3847, 2023.

\bibitem{blattmann2023stable}
Andreas Blattmann, Tim Dockhorn, Sumith Kulal, Daniel Mendelevitch, Maciej
  Kilian, Dominik Lorenz, Yam Levi, Zion English, Vikram Voleti, Adam Letts,
  et~al.
\newblock Stable video diffusion: Scaling latent video diffusion models to
  large datasets.
\newblock {\em arXiv preprint arXiv:2311.15127}, 2023.

\bibitem{poole2022dreamfusion}
Ben Poole, Ajay Jain, Jonathan~T Barron, and Ben Mildenhall.
\newblock Dreamfusion: Text-to-3d using 2d diffusion.
\newblock {\em arXiv preprint arXiv:2209.14988}, 2022.

\bibitem{wang2023score}
Haochen Wang, Xiaodan Du, Jiahao Li, Raymond~A Yeh, and Greg Shakhnarovich.
\newblock Score jacobian chaining: Lifting pretrained 2d diffusion models for
  3d generation.
\newblock In {\em Proceedings of the IEEE/CVF conference on computer vision and
  pattern recognition}, pages 12619--12629, 2023.

\bibitem{wang2019planit}
Kai Wang, Yu-An Lin, Ben Weissmann, Manolis Savva, Angel~X Chang, and Daniel
  Ritchie.
\newblock Planit: Planning and instantiating indoor scenes with relation graph
  and spatial prior networks.
\newblock {\em ACM Transactions on Graphics (TOG)}, 38(4):1--15, 2019.

\bibitem{dhamo2021graph}
Helisa Dhamo, Fabian Manhardt, Nassir Navab, and Federico Tombari.
\newblock Graph-to-3d: End-to-end generation and manipulation of 3d scenes
  using scene graphs.
\newblock In {\em Proceedings of the IEEE/CVF International Conference on
  Computer Vision}, pages 16352--16361, 2021.

\bibitem{zhai2023commonscenes}
Guangyao Zhai, Evin~P{\i}nar {\"O}rnek, Shun-Cheng Wu, Yan Di, Federico
  Tombari, Nassir Navab, and Benjamin Busam.
\newblock Commonscenes: Generating commonsense 3d indoor scenes with scene
  graph diffusion.
\newblock {\em Advances in Neural Information Processing Systems},
  36:30026--30038, 2023.

\bibitem{lin2024instructscene}
Chenguo Lin and Yadong Mu.
\newblock Instructscene: Instruction-driven 3d indoor scene synthesis with
  semantic graph prior.
\newblock {\em arXiv preprint arXiv:2402.04717}, 2024.

\bibitem{zhai2024echoscene}
Guangyao Zhai, Evin~P{\i}nar {\"O}rnek, Dave~Zhenyu Chen, Ruotong Liao, Yan Di,
  Nassir Navab, Federico Tombari, and Benjamin Busam.
\newblock Echoscene: Indoor scene generation via information echo over scene
  graph diffusion.
\newblock In {\em European Conference on Computer Vision}, pages 167--184.
  Springer, 2024.

\bibitem{li2019grains}
Manyi Li, Akshay~Gadi Patil, Kai Xu, Siddhartha Chaudhuri, Owais Khan, Ariel
  Shamir, Changhe Tu, Baoquan Chen, Daniel Cohen-Or, and Hao Zhang.
\newblock Grains: Generative recursive autoencoders for indoor scenes.
\newblock {\em ACM Transactions on Graphics (TOG)}, 38(2):1--16, 2019.

\bibitem{bahmani2023cc3d}
Sherwin Bahmani, Jeong~Joon Park, Despoina Paschalidou, Xingguang Yan, Gordon
  Wetzstein, Leonidas Guibas, and Andrea Tagliasacchi.
\newblock Cc3d: Layout-conditioned generation of compositional 3d scenes.
\newblock In {\em Proceedings of the IEEE/CVF International Conference on
  Computer Vision}, pages 7171--7181, 2023.

\bibitem{po2024compositional}
Ryan Po and Gordon Wetzstein.
\newblock Compositional 3d scene generation using locally conditioned
  diffusion.
\newblock In {\em 2024 International Conference on 3D Vision (3DV)}, pages
  651--663. IEEE, 2024.

\bibitem{yang2024scenecraft}
Xiuyu Yang, Yunze Man, Junkun Chen, and Yu-Xiong Wang.
\newblock Scenecraft: Layout-guided 3d scene generation.
\newblock {\em Advances in Neural Information Processing Systems},
  37:82060--82084, 2024.

\bibitem{chen2025layout2scene}
Minglin Chen, Longguang Wang, Sheng Ao, Ye~Zhang, Kai Xu, and Yulan Guo.
\newblock Layout2scene: 3d semantic layout guided scene generation via geometry
  and appearance diffusion priors.
\newblock {\em arXiv preprint arXiv:2501.02519}, 2025.

\bibitem{chen2023fantasia3d}
Rui Chen, Yongwei Chen, Ningxin Jiao, and Kui Jia.
\newblock Fantasia3d: Disentangling geometry and appearance for high-quality
  text-to-3d content creation.
\newblock In {\em Proceedings of the IEEE/CVF international conference on
  computer vision}, pages 22246--22256, 2023.

\bibitem{wang2023prolificdreamer}
Zhengyi Wang, Cheng Lu, Yikai Wang, Fan Bao, Chongxuan Li, Hang Su, and Jun
  Zhu.
\newblock Prolificdreamer: High-fidelity and diverse text-to-3d generation with
  variational score distillation.
\newblock {\em Advances in neural information processing systems},
  36:8406--8441, 2023.

\bibitem{lin2023magic3d}
Chen-Hsuan Lin, Jun Gao, Luming Tang, Towaki Takikawa, Xiaohui Zeng, Xun Huang,
  Karsten Kreis, Sanja Fidler, Ming-Yu Liu, and Tsung-Yi Lin.
\newblock Magic3d: High-resolution text-to-3d content creation.
\newblock In {\em Proceedings of the IEEE/CVF conference on computer vision and
  pattern recognition}, pages 300--309, 2023.

\bibitem{yi2024gaussiandreamer}
Taoran Yi, Jiemin Fang, Junjie Wang, Guanjun Wu, Lingxi Xie, Xiaopeng Zhang,
  Wenyu Liu, Qi~Tian, and Xinggang Wang.
\newblock Gaussiandreamer: Fast generation from text to 3d gaussians by
  bridging 2d and 3d diffusion models.
\newblock In {\em Proceedings of the IEEE/CVF conference on computer vision and
  pattern recognition}, pages 6796--6807, 2024.

\bibitem{wang2025vggt}
Jianyuan Wang, Minghao Chen, Nikita Karaev, Andrea Vedaldi, Christian
  Rupprecht, and David Novotny.
\newblock Vggt: Visual geometry grounded transformer.
\newblock In {\em Proceedings of the Computer Vision and Pattern Recognition
  Conference}, pages 5294--5306, 2025.

\bibitem{hong20233d}
Yining Hong, Haoyu Zhen, Peihao Chen, Shuhong Zheng, Yilun Du, Zhenfang Chen,
  and Chuang Gan.
\newblock 3d-llm: Injecting the 3d world into large language models.
\newblock {\em Advances in Neural Information Processing Systems},
  36:20482--20494, 2023.

\bibitem{xu2024pointllm}
Runsen Xu, Xiaolong Wang, Tai Wang, Yilun Chen, Jiangmiao Pang, and Dahua Lin.
\newblock Pointllm: Empowering large language models to understand point
  clouds.
\newblock In {\em European Conference on Computer Vision}, pages 131--147.
  Springer, 2024.

\bibitem{he2016deep}
Kaiming He, Xiangyu Zhang, Shaoqing Ren, and Jian Sun.
\newblock Deep residual learning for image recognition.
\newblock In {\em Proceedings of the IEEE conference on computer vision and
  pattern recognition}, pages 770--778, 2016.

\bibitem{radford2021learning}
Alec Radford, Jong~Wook Kim, Chris Hallacy, Aditya Ramesh, Gabriel Goh,
  Sandhini Agarwal, Girish Sastry, Amanda Askell, Pamela Mishkin, Jack Clark,
  et~al.
\newblock Learning transferable visual models from natural language
  supervision.
\newblock In {\em International conference on machine learning}, pages
  8748--8763. PmLR, 2021.

\bibitem{kerbl20233d}
Bernhard Kerbl, Georgios Kopanas, Thomas Leimk{\"u}hler, George Drettakis,
  et~al.
\newblock 3d gaussian splatting for real-time radiance field rendering.
\newblock {\em ACM Trans. Graph.}, 42(4):139--1, 2023.

\bibitem{worldlabs2025marble}
{World Labs}.
\newblock Marble: A multimodal world model.
\newblock \url{https://www.worldlabs.ai/blog/marble-world-model}, November
  2025.
\newblock Accessed: 2026-03-02.

\bibitem{chang2015shapenet}
Angel~X Chang, Thomas Funkhouser, Leonidas Guibas, Pat Hanrahan, Qixing Huang,
  Zimo Li, Silvio Savarese, Manolis Savva, Shuran Song, Hao Su, et~al.
\newblock Shapenet: An information-rich 3d model repository.
\newblock {\em arXiv preprint arXiv:1512.03012}, 2015.

\bibitem{deitke2023objaverse}
Matt Deitke, Ruoshi Liu, Matthew Wallingford, Huong Ngo, Oscar Michel, Aditya
  Kusupati, Alan Fan, Christian Laforte, Vikram Voleti, Samir~Yitzhak Gadre,
  et~al.
\newblock Objaverse-xl: A universe of 10m+ 3d objects.
\newblock {\em Advances in Neural Information Processing Systems},
  36:35799--35813, 2023.

\bibitem{wu2023omniobject3d}
Tong Wu, Jiarui Zhang, Xiao Fu, Yuxin Wang, Jiawei Ren, Liang Pan, Wayne Wu,
  Lei Yang, Jiaqi Wang, Chen Qian, et~al.
\newblock Omniobject3d: Large-vocabulary 3d object dataset for realistic
  perception, reconstruction and generation.
\newblock In {\em Proceedings of the IEEE/CVF conference on computer vision and
  pattern recognition}, pages 803--814, 2023.

\bibitem{sun2018pix3d}
Xingyuan Sun, Jiajun Wu, Xiuming Zhang, Zhoutong Zhang, Chengkai Zhang, Tianfan
  Xue, Joshua~B Tenenbaum, and William~T Freeman.
\newblock Pix3d: Dataset and methods for single-image 3d shape modeling.
\newblock In {\em Proceedings of the IEEE conference on computer vision and
  pattern recognition}, pages 2974--2983, 2018.

\bibitem{collins2022abo}
Jasmine Collins, Shubham Goel, Kenan Deng, Achleshwar Luthra, Leon Xu, Erhan
  Gundogdu, Xi~Zhang, Tomas F~Yago Vicente, Thomas Dideriksen, Himanshu Arora,
  et~al.
\newblock Abo: Dataset and benchmarks for real-world 3d object understanding.
\newblock In {\em Proceedings of the IEEE/CVF conference on computer vision and
  pattern recognition}, pages 21126--21136, 2022.

\bibitem{wu20153d}
Zhirong Wu, Shuran Song, Aditya Khosla, Fisher Yu, Linguang Zhang, Xiaoou Tang,
  and Jianxiong Xiao.
\newblock 3d shapenets: A deep representation for volumetric shapes.
\newblock In {\em Proceedings of the IEEE conference on computer vision and
  pattern recognition}, pages 1912--1920, 2015.

\bibitem{dai2017scannet}
Angela Dai, Angel~X Chang, Manolis Savva, Maciej Halber, Thomas Funkhouser, and
  Matthias Nie{\ss}ner.
\newblock Scannet: Richly-annotated 3d reconstructions of indoor scenes.
\newblock In {\em Proceedings of the IEEE conference on computer vision and
  pattern recognition}, pages 5828--5839, 2017.

\bibitem{fu20213d}
Huan Fu, Bowen Cai, Lin Gao, Ling-Xiao Zhang, Jiaming Wang, Cao Li, Qixun Zeng,
  Chengyue Sun, Rongfei Jia, Binqiang Zhao, et~al.
\newblock 3d-front: 3d furnished rooms with layouts and semantics.
\newblock In {\em Proceedings of the IEEE/CVF International Conference on
  Computer Vision}, pages 10933--10942, 2021.

\bibitem{he2023t}
Yuze He, Yushi Bai, Matthieu Lin, Wang Zhao, Yubin Hu, Jenny Sheng, Ran Yi,
  Juanzi Li, and Yong-Jin Liu.
\newblock T3bench: Benchmarking current progress in text-to-3d generation.
\newblock {\em arXiv preprint arXiv:2310.02977}, 2023.

\bibitem{chan2022efficient}
Eric~R Chan, Connor~Z Lin, Matthew~A Chan, Koki Nagano, Boxiao Pan, Shalini
  De~Mello, Orazio Gallo, Leonidas~J Guibas, Jonathan Tremblay, Sameh Khamis,
  et~al.
\newblock Efficient geometry-aware 3d generative adversarial networks.
\newblock In {\em Proceedings of the IEEE/CVF conference on computer vision and
  pattern recognition}, pages 16123--16133, 2022.

\bibitem{gao2022get3d}
Jun Gao, Tianchang Shen, Zian Wang, Wenzheng Chen, Kangxue Yin, Daiqing Li,
  Or~Litany, Zan Gojcic, and Sanja Fidler.
\newblock Get3d: A generative model of high quality 3d textured shapes learned
  from images.
\newblock {\em Advances in neural information processing systems},
  35:31841--31854, 2022.

\bibitem{muller2023diffrf}
Norman M{\"u}ller, Yawar Siddiqui, Lorenzo Porzi, Samuel~Rota Bulo, Peter
  Kontschieder, and Matthias Nie{\ss}ner.
\newblock Diffrf: Rendering-guided 3d radiance field diffusion.
\newblock In {\em Proceedings of the IEEE/CVF conference on computer vision and
  pattern recognition}, pages 4328--4338, 2023.

\bibitem{cao2023large}
Ziang Cao, Fangzhou Hong, Tong Wu, Liang Pan, and Ziwei Liu.
\newblock Large-vocabulary 3d diffusion model with transformer.
\newblock {\em arXiv preprint arXiv:2309.07920}, 2023.

\bibitem{zhou2024diffgs}
Junsheng Zhou, Weiqi Zhang, and Yu-Shen Liu.
\newblock Diffgs: Functional gaussian splatting diffusion.
\newblock {\em Advances in Neural Information Processing Systems},
  37:37535--37560, 2024.

\bibitem{lan2024ln3diff}
Yushi Lan, Fangzhou Hong, Shuai Yang, Shangchen Zhou, Xuyi Meng, Bo~Dai,
  Xingang Pan, and Chen~Change Loy.
\newblock Ln3diff: Scalable latent neural fields diffusion for speedy 3d
  generation.
\newblock In {\em ECCV}, 2024.

\bibitem{chung2023luciddreamer}
Jaeyoung Chung, Suyoung Lee, Hyeongjin Nam, Jaerin Lee, and Kyoung~Mu Lee.
\newblock Luciddreamer: Domain-free generation of 3d gaussian splatting scenes.
\newblock {\em arXiv preprint arXiv:2311.13384}, 2023.

\bibitem{chen2024single}
Yixin Chen, Junfeng Ni, Nan Jiang, Yaowei Zhang, Yixin Zhu, and Siyuan Huang.
\newblock Single-view 3d scene reconstruction with high-fidelity shape and
  texture.
\newblock In {\em 2024 International Conference on 3D Vision (3DV)}, pages
  1456--1467. IEEE, 2024.

\bibitem{ardelean2025gen3dsr}
Andreea Ardelean, Mert {\"O}zer, and Bernhard Egger.
\newblock Gen3dsr: Generalizable 3d scene reconstruction via divide and conquer
  from a single view.
\newblock In {\em 2025 International Conference on 3D Vision (3DV)}, pages
  616--626. IEEE, 2025.

\bibitem{han2025reparo}
Haonan Han, Rui Yang, Huan Liao, Jiankai Xing, Zunnan Xu, Xiaoming Yu, Junwei
  Zha, Xiu Li, and Wanhua Li.
\newblock Reparo: Compositional 3d assets generation with differentiable 3d
  layout alignment.
\newblock In {\em Proceedings of the IEEE/CVF International Conference on
  Computer Vision}, pages 25367--25377, 2025.

\bibitem{huang2025midi}
Zehuan Huang, Yuan-Chen Guo, Xingqiao An, Yunhan Yang, Yangguang Li, Zi-Xin
  Zou, Ding Liang, Xihui Liu, Yan-Pei Cao, and Lu~Sheng.
\newblock Midi: Multi-instance diffusion for single image to 3d scene
  generation.
\newblock In {\em Proceedings of the IEEE/CVF Conference on Computer Vision and
  Pattern Recognition}, pages 23646--23657, 2025.

\bibitem{wu20244d}
Guanjun Wu, Taoran Yi, Jiemin Fang, Lingxi Xie, Xiaopeng Zhang, Wei Wei, Wenyu
  Liu, Qi~Tian, and Xinggang Wang.
\newblock 4d gaussian splatting for real-time dynamic scene rendering.
\newblock In {\em Proceedings of the IEEE/CVF conference on computer vision and
  pattern recognition}, pages 20310--20320, 2024.

\end{thebibliography}
\end{document}